\documentclass{article} % For LaTeX2e
\usepackage{iclr2020_conference,times}
\usepackage{textcomp}

\usepackage{amsmath,amsfonts,bm}

\def\eqref#1{equation~\ref{#1}}
\def\1{\bm{1}}

\DeclareMathAlphabet{\mathsfit}{\encodingdefault}{\sfdefault}{m}{sl}
\SetMathAlphabet{\mathsfit}{bold}{\encodingdefault}{\sfdefault}{bx}{n}

\usepackage{graphicx}
\usepackage{hyperref}
\usepackage{url}
\usepackage{multirow}
\usepackage{xcolor}
\usepackage[T4,T1]{fontenc}
\newenvironment{tfour}{\fontencoding{T4}\selectfont}{}
\usepackage{newunicodechar}
\usepackage{subcaption}

\newunicodechar{ɔ}{\fon}

\usepackage{tabularx}
\title{FFR v1.0 : Fon-French Neural Machine Translation 
}
\author{Bonaventure F. P. Dossou \& Chris C. Emezue\\
Department of Theories of Functions and Approximations\\
Kazan (Volga) Federal University\\
Kazan, Russian Federation\\
\texttt{\{femipancrace.dossou,chris.emezue\}@gmail.com} \\
}

\iclrfinalcopy % Uncomment for camera-ready version, but NOT for submission.
\begin{document}

\maketitle

\section{Introduction}
Africa has the highest linguistic diversity in the world\citep{africa}. On account of the importance of language to communication, and the importance of reliable, powerful and accurate machine translation models in modern inter-cultural communication, there have been (and still are) efforts to create state-of-the-art translation models for the many African languages. However, the low-resources, diacritical and tonal complexities of African languages are major issues facing African NLP today. The FFR is a major step towards creating a robust translation model from Fon, a very low-resource and tonal language, to French, for research and public use. In this paper, we describe our pilot project: the creation of a large growing corpora for Fon-to-French translations and our FFR v1.0 model, trained on this dataset. The dataset and model are made publicly available.
\section{Motivation}
The selection of Fon language for this pilot project is guided by the fact that not only is Fon one of the major languages spoken by the natives of Benin Republic, it belongs to, and shares tonal and analytical similarities with the Niger-Congo languages, which is the largest group of African languages comprising widely spoken languages like Igbo, Hausa, Yoruba and Swahili \citep{greenberg}. French was chosen because it is the European language spoken officially by the natives of Fon and both languages contain diacritics. Therefore, a machine translation model that succeeds for the Fon-French can be trained (with transfer learning) on the other large group of Niger-Congo African languages and European languages.
\section{Related works}
It is important to note that the move to promote African NLP has been going on for quite a while, with the advent of organizations and online communities  like \href{https://twitter.com/DeepIndaba}{Deep Learning Indaba}, \href{https://blackinai.github.io/}{BlackinAI}, and most inspirational to this paper, \href{https://www.masakhane.io/home}{Masakhane}, an online community focused on connecting and fostering machine translation (MT) researchers on African languages who have all made meaningful contributions to MT on some African languages. \href{https://github.com/kevindegila}{Kevin Degila}, a member of the community worked on an English-Fon translation model.
\section{Project FFR v1.0}
\subsection{FFR Dataset}
We created the FFR Dataset as a project to compile a large, growing corpora of cleaned Fon - French sentences for machine translation, and other NLP research-related, projects. As training data is crucial to the high performance of a machine learning model, we hope to facilitate future research being done on the Fon language, by releasing our data for research purposes. The major sources for the creation of FFR Dataset were:\\
\begin{enumerate}
\item JW300 - \url{http://opus.nlpl.eu/JW300.php} (24.60\% of FFR Dataset)
\item BeninLanguages - \url{https://beninlangues.com/}  (75.40\% of FFR Dataset)
\end{enumerate}

JW300 is a parallel corpus of over 300 languages with around 100 thousand parallel sentences per language pair on average. BeninLanguages contains (in French and Fon) vocabulary words, short expressions, small sentences, complex sentences, proverbs and bible verses: Genesis 1 - Psalm 79.

The tabular analysis shown in Table~\ref{ffrdataset} below serves to give an idea of the range of word lengths for the sentences in the FFR dataset. The maximum number of words for the fon sentences, $max-fon$, is $109$, while that of the french sentences, $max-fr$, is $111$. This shows that the FFR Dataset is mostly made up of very short sentences containing 1-5 words, but at the same time, contains some medium to long sentences, thus achieving the intended variety of the dataset.

\begin{table}[h]
\caption{ \bf Analysis of sentences in the FFR dataset}
\label{ffrdataset}
\begin{center}
\begin{tabular}{lll}
\multicolumn{1}{l}{\bf \#} &\multicolumn{1}{l}{\bf Fon} & \multicolumn{1}{l}{\bf French} \\
\\ \hline \\
Very Short sentences (1-5 words)     &64301          &64255        \\
Short sentences (6-10 words)            &13848                    &17183   \\
Medium sentences (11-30 words)             &29113    &29857 \\
Long sentences (31+ words)         &9767           &5734 \\
\end{tabular}
\end{center}
\end{table}

The FFR Dataset and the official documentation, with more information on the dataset, which could not be detailed here due to the page-limit, can be found at \url{https://github.com/bonaventuredossou/ffr-v1/tree/master/FFR-Dataset}.  The FFR Dataset currently contains 117,029 parallel Fon-French words and sentences, which we used to train the FFR v1.0 model.
\subsection{FFR v1.0 Model}
\subsubsection{Data Preprocessing}
Our earlier effort at pre-analysis of Fon sentences, which revealed that different accents on same words affected their meanings as shown in sentences \#2 and \#3 of Table~\ref{sentences-table}, made it necessary to also design our own strategy for encoding the diacritics of the Fon languages thus: we encoded the words with their different accents, instead of the default, which removed all accents.

\begin{table}[h]
\caption{ \bf{Number of samples (sentences) for training FFR}}
\label{dataset}
\begin{center}
\begin{tabular}{ll}
\multicolumn{1}{c}{}  &\multicolumn{1}{c}{} 
\\ \hline \\
\textbf{Training}      &105,326 \\
\textbf{Validation}             &5,691 \\
\textbf{Testing}             &6,012 \\
\end{tabular}
\end{center}
\end{table}

\subsubsection{Model structure}
In summary, the FFR model is based on the encoder-decoder configuration\citep{brownlee,tensorflow}. Our encoders and decoders are made up of 128-dimensional gated rectified units (GRUs) recurrent layers, with a word embedding layer of dimension 512. The encoder transforms a source sentence to a fixed-length context vector and the decoder learns to interpret the output from the encoded vector by maximizing the probability of a correct translation given a source sentence. We also applied a 30-dimensional attention model\citep{NIPS20145346,attention,lamba} in order to help the model learn which words to place attention on and make contextual, correct translations. The code for the model has been \href{https://github.com/bonaventuredossou/ffr-v1/blob/master/model_train_test/fon_fr.py}{open-sourced on GitHub}.
\subsubsection{Initial Results and Findings}

\begin{table}[h]
\caption{Overall BLEU and GLUE scores on FFR v1.0 test samples}
\label{scores}
\begin{center}
\begin{tabular}{lll}
\multicolumn{1}{c}{\bf Model Structure}  &\multicolumn{1}{c}{\bf BLEU}  &\multicolumn{1}{c}{\bf GLUE}
\\ \hline \\
without diacritical encoding         &24.53 & 13.0 \\
with diacritical encoding             &\textbf{30.55} &\textbf{18.18} \\
\end{tabular}
\end{center}
\end{table}

We trained the model with and without diacritical encoding. As seen in Table~\ref{scores}, our diacritical encoding greatly improved the performance of the FFR model on both the BLEU and GLUE (a modification of the BLEU metric introduced my Google) metrics. This really shows how important diacritics are in our african language structures and meaning and therefore the need to build models that can interprete them very well.

\begin{table}[h]
\caption{Sentences predictions and scores}
\label{sentences-table}
\begin{center}
 \resizebox{\textwidth}{!}{
\begin{tabular}{ |c|c|c|c|c|c|c| } 
 \hline
\bf ID & \bf 0 &\bf 1 &\bf2&\bf3&\bf4&\bf5\\
 \textcolor{blue}{Source} & \textcolor{blue}{yí bo wa} & \textcolor{blue}{yi bo wa}  &\textcolor{blue}{h(\')\begin{tfour}\m{o}\end{tfour}n} &\textcolor{blue}{h\begin{tfour}\m{o}\end{tfour}n} & \textcolor{blue}{s\'a amas\'in d\u o w\u u } &\textcolor{blue}{gb(\')\begin{tfour}\m{e}\end{tfour}}
 \\ 
 \textcolor{brown}{Target} &  \textcolor{brown}{prends et viens} &  \textcolor{brown}{va et viens} & \textcolor{brown}{porte} & \textcolor{brown}{fuire} & \textcolor{brown}{oindre avec un m\'edicament} &\textcolor{brown}{pousser de nouvelles feuilles}  \\
 FFR v1.0 Model & prends et viens & va viens & scorpion &porte &se masser le remede &esprit de la vie\\
 BLEU/\bf CMS score & 1.0 & 1.0 & 0.0   &0.0 &0.0/\textbf{0.65} &0.25 /\textbf{ 0.9} \\ 
 \hline
\end{tabular} 
}
\end{center}
\end{table}

Table~\ref{sentences-table} shows translations of interest from the FFR model, illustrating the difficulty of predicting Fon words which bear different meanings with different accents. While our model predicted well for \#0 and \#1, it misplaced the meanings for \#2 and \#3.  

\textbf{CMS(context-meaning-similarity):} We discovered that the FFR v1.0 model was able to provide predictions that were, although different from the target, similar in context to the target, as seen in sentence \#4. This led us to develop the CMS metric: we sent the source and target sentences to five Fon-French natives, and requested a score from 0 to 1 on how similar in contextual meaning the predictions were with the source sentences and target sentences. Then we took the average of their reviews as the CMS score for each of the model\textquotesingle s predictions as given in sentence \#4 in Table~\ref{sentences-table}. 

We also discovered that there were cases, for example sentence \#5, where the target was wrong, but the model was able to predict a correct translation.

Although the CMS metric is crude at the moment, we believe there is potential in exploring it further: it could shed more light on measuring the performance of a translation model for these tonal African languages, as well as out-of-vocabulary translation performance.

The FFR model is a pilot project and there is headroom to be explored with the tuning of different architectures, learning schemes, and transfer learning for FFR Model v2.0.

\bibliography{iclr2020_conference}
\bibliographystyle{iclr2020_conference}

\end{document}